%% file: sample-authordraft.tex
\documentclass[sigconf,natbib=true]{acmart}
\makeatletter
\def\@ACM@checkaffil{
    \if@ACM@instpresent\else
    \ClassWarningNoLine{\@classname}{No institution present for an affiliation}%
    \fi
    \if@ACM@citypresent\else
    \ClassWarningNoLine{\@classname}{No city present for an affiliation}%
    \fi
    \if@ACM@countrypresent\else
        \ClassWarningNoLine{\@classname}{No country present for an affiliation}%
    \fi
}
\makeatother
\usepackage{algorithm}
\usepackage{microtype}
\usepackage{algcompatible,amsmath}
\usepackage{graphicx}
\usepackage{url}
\usepackage{multirow}
\usepackage{comment}
\usepackage{pifont}
\usepackage{balance}
\usepackage{mathtools}
\usepackage{xcolor}
\usepackage{colortbl}
\usepackage{bbm}
\usepackage{enumitem}
\usepackage{marvosym}
\usepackage{textcomp}
\usepackage{array}
\usepackage{booktabs}
\usepackage{tabularx}
\usepackage{CJKutf8}
\usepackage{algorithm}  
\usepackage{algorithmicx}  
\usepackage{algpseudocode} 

\usepackage{fancyhdr}
\usepackage{newfloat}
\usepackage{listings}
\floatstyle{ruled}
\newfloat{listing}{tb}{lst}{}
\floatname{listing}{Listing}

\newcommand{\squishlist}{
\begin{list}{$\bullet$}
{ \usecounter{Lcount}
\setlength{\itemsep}{0pt}
\setlength{\parsep}{0pt}
\setlength{\topsep}{0pt}
\setlength{\partopsep}{0pt}
\setlength{\leftmargin}{2em}
\setlength{\labelwidth}{1.5em}
\setlength{\labelsep}{0.5em} } }
\newcommand{\squishend}{
\end{list} }
\newcommand*\circled[1]{\kern-2.5em%
  \put(0,4){\color{white}\circle*{18}}\put(0,4){\circle{10}}%
  \put(-3,0){\color{black}\bfseries#1}~~}
\usepackage{tikz} 

\usepackage{enumitem}
\newcommand{\modelname}{{\textsc{\textsf{RE$^{2}$}}}}
\makeatletter
\newcommand{\thickhline}{%
    \noalign {\ifnum 0=`}\fi \hrule height 2pt
    \futurelet \reserved@a \@xhline
}

\definecolor{codegreen}{rgb}{0,0.6,0}
\definecolor{codegray}{rgb}{0.5,0.5,0.5}
\definecolor{codepurple}{rgb}{0.58,0,0.82}
\definecolor{backcolour}{rgb}{0.95,0.95,0.92}

\lstdefinestyle{mystyle}{
    backgroundcolor=\color{backcolour},   
    commentstyle=\color{codegreen},
    keywordstyle=\color{magenta},
    numberstyle=\tiny\color{codegray},
    stringstyle=\color{codepurple},
    basicstyle=\ttfamily\footnotesize,
    breakatwhitespace=false,         
    breaklines=true,                 
    captionpos=b,                    
    keepspaces=true,                 
    numbers=left,                    
    numbersep=5pt,                  
    showspaces=false,                
    showstringspaces=false,
    showtabs=false,                  
    tabsize=2
}

\copyrightyear{2023}
\acmYear{2023}
\setcopyright{rightsretained}

\acmConference[SIGIR '23]{Proceedings of the 46th International ACM SIGIR Conference on Research and Development in Information Retrieval}
{July 23--27, 2023}
{Taipei, Taiwan}
\acmBooktitle{Proceedings of the 46th International ACM SIGIR Conference on Research and Development in Information Retrieval (SIGIR '23), July 23--27, 2023, Taipei, Taiwan}
\acmDOI{10.1145/3539618.3592072}
\acmISBN{978-1-4503-9408-6/23/07}

\usepackage{etoolbox}
\makeatletter
\patchcmd{\maketitle}{\@copyrightpermission}{
   \begin{minipage}{0.3\columnwidth}
     \href{http://creativecommons.org/licenses/by/4.0/}{\includegraphics[width=0.90\textwidth]{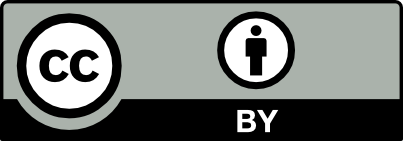}}
   \end{minipage}\hfill
   \begin{minipage}{0.7\columnwidth}
     \href{http://creativecommons.org/licenses/by/4.0/}{This work is licensed under a Creative Commons Attribution International 4.0 License.}
   \end{minipage}
}{}{}

\makeatother

\begin{document}

\title{Think Rationally about What You See: Continuous Rationale Extraction for Relation Extraction}

\author{Xuming Hu}
\affiliation{%
  \institution{Tsinghua University}
}
\email{hxm19@mails.tsinghua.edu.cn}

\author{Zhaochen Hong}
\affiliation{%
  \institution{Tsinghua University}
}
\email{hongzc20@mails.tsinghua.edu.cn}

\author{Chenwei Zhang}

\affiliation{%
  \institution{Amazon}
  }
\email{cwzhang910@gmail.com}

\author{Irwin King}

\affiliation{%
 \institution{The Chinese University of Hong Kong} %
 }
\email{king@cse.cuhk.edu.hk}

\author{Philip S. Yu}
\affiliation{%
 \institution{University of Illinois at Chicago}
 }
\email{psyu@cs.uic.edu}
\renewcommand{\shortauthors}{Hu, et al.}


\begin{abstract}
Relation extraction (RE) aims to extract potential relations according to the context of two entities, thus, deriving rational contexts from sentences plays an important role. Previous works either focus on how to leverage the entity information (e.g., entity types, entity verbalization) to inference relations, but ignore context-focused content, or use counterfactual thinking to remove the model's bias of potential relations in entities, but the relation reasoning process will still be hindered by irrelevant content. Therefore, how to preserve relevant content and remove noisy segments from sentences is a crucial task. In addition, retained content needs to be fluent enough to maintain semantic coherence and interpretability. In this work, we propose a novel rationale extraction framework named {\modelname}, which leverages two continuity and sparsity factors to obtain relevant and coherent rationales from sentences. 
To solve the problem that the gold rationales are not labeled, {\modelname} applies an optimizable binary mask to each token in the sentence, and adjust the rationales that need to be selected according to the relation label. Experiments on four datasets show that {\modelname} surpasses baselines. 
\end{abstract}







\maketitle
\input{subfiles/1_intro}

\input{subfiles/4_model}

\input{subfiles/5_experiment}
\input{subfiles/6_conclusion}

\section{Acknowledgments}
Chenwei Zhang is the corresponding author. Xuming Hu and Zhaochen Hong have made equal contributions to this work. Source code is available at: \url{https://github.com/THU-BPM/RE2}. The work described in this paper was partially supported by the National Key Research and Development Program of China (No. 2018AAA0100204) and by the Research Grants Council of the Hong Kong Special Administrative Region, China (RGC GRF 2151185; CUHK 14222922), NSF under grant III-1909323.

\balance
\bibliographystyle{ACM-Reference-Format}
\bibliography{sample-base}

\appendix
\newpage









\end{document}

%% file: subfiles/1_intro.tex
\section{Introduction}
\label{intro}
\begin{figure}
    \centering
    \includegraphics[scale=0.32]{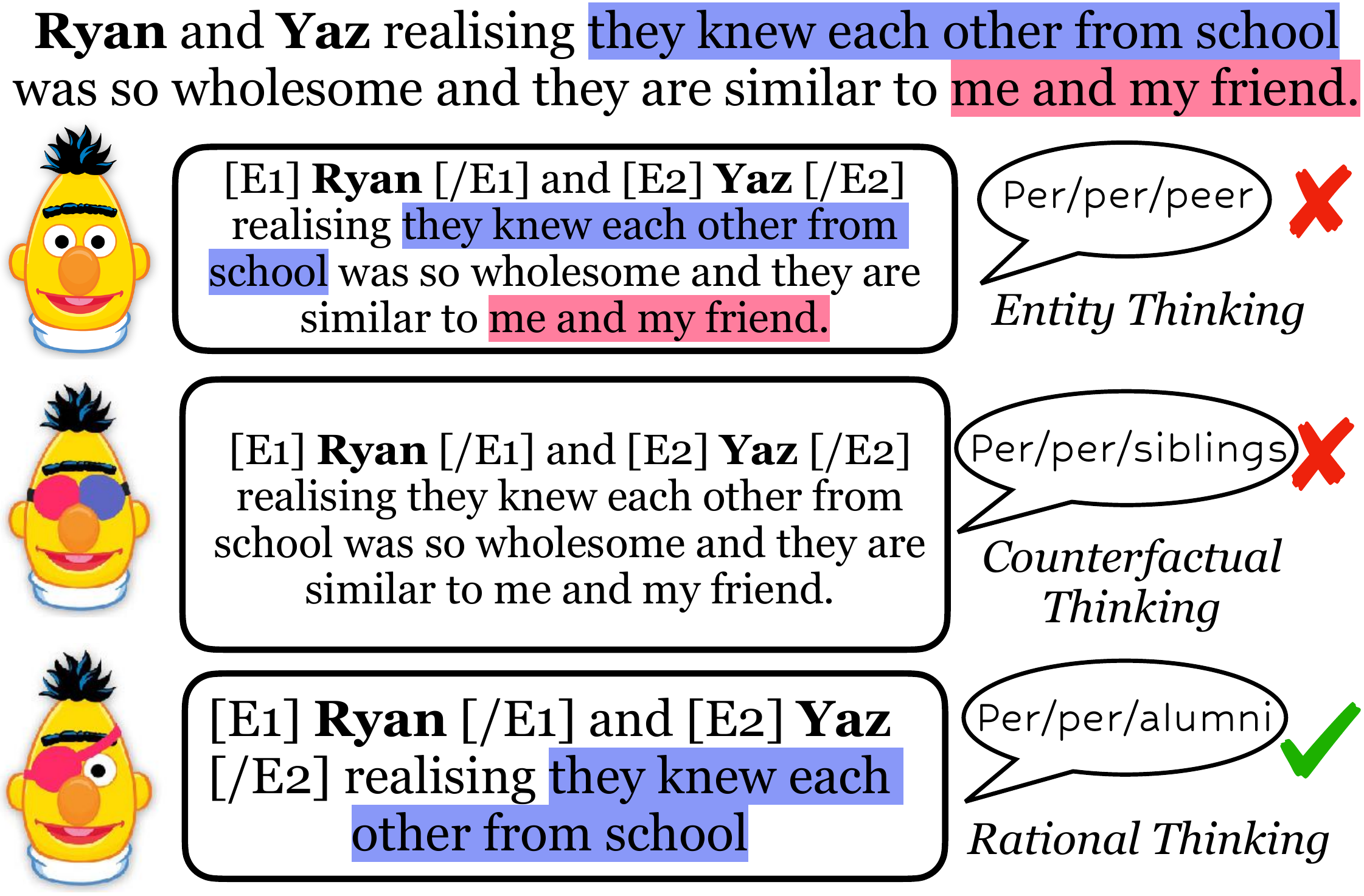}
    \vspace{-3mm}
    \caption{Different models ``see'' different content in sentences by thinking differently. Rational thinking predicts the correct relation label \texttt{per/per/alumni} between two entities \texttt{Ryan} and \texttt{Yaz} by seeing the relevant and correct content.}
    \label{fig:example}
    \vspace{-5mm}
\end{figure}

Relation extraction (RE) is a crucial part of many information retrieval (IR) systems, which could extract relations between entities from sentences. These structured triplets such as \textit{(Ryan, Yaz, per/per/alumni)} (Figure \ref{fig:example}) from heterogeneous sources could benefit multiple downstream applications like question answering \cite{liu2022semantic,liu2023comprehensive} and natural language understanding \cite{liu2022hierarchical,hu2022scene,liu2022character}. To obtain structured triples, we need to exploit relevant and noise-free sentences from the entity context, so that the correct relation can be extracted between the two available entities. \textit{Entity Thinking} methods such as \citet{hu2020selfore,hu2021semi,hu2021gradient} inject reserved special tokens <e> and </e> before and after the entity, and focus on the contextualized features of the entity through these special tokens. However, the semantic information of entities cannot be specified in special tokens. Therefore, \citet{zhou2022improved} and \citet{lu2022summarization} respectively introduce entity type and entity verbalization to better reveal the contextual semantic representation of entities and infer the relations between entities. Although Entity Thinking methods can better capture the context semantics of the entity, but cannot automatically remove noisy and irrelevant contents. Such noisy content tends to destroy correct relational inference. Taking Figure \ref{fig:example} as an example, the Entity Thinking method has no idea in judging the relevance of the content such as ``they knew each other from school'' and ``me and my friend'' for predicting the relation between entities. Therefore, the model may be misled by the word ``friend'', and mispredicts the relation as \texttt{Per/per/peer}. To remove the potential impact of the content on the relation extraction between entities, \textit{Counterfactual Thinking} methods \cite{nan2021uncovering,wang2022should} remove the model's bias against different words and entities. 
However, these methods do not focus on explicitly removing noisy contextual content, thus, the model's prediction can still be misled as \texttt{Per/per/siblings}.

To remove irrelevant and noisy content in sentences, we first propose rational thinking methods which could extract relevant and noise-free rationales in RE task. Although the methods of rational thinking have been verified in the various downstream tasks of information retrieval such as question answering \cite{zhang2018deep}, we still face two crucial challenges to leverage the rational thinking methods for the RE task: (1) Gold rationales which are relevant to the relation label in the sentences are not available, therefore, we cannot train a rationale extractor in a supervised learning manner, (2) Extracted rationales are encouraged to be continuous, which can not only improve the interpretability of the rationales, but also express coherent semantics to predict the relation labels between entities. 

In addressing the two main challenges, we present two new continuity and sparsity factors in this study to manage the coherence and quantity of chosen rationale tokens. Sparsity imposition aids in striking a balance between eliminating irrelevant material and preserving relevant content. Promoting continuity is advantageous for obtaining continuous rationales, leading to a more coherent semantic representation. Furthermore, we employ an adjustable binary mask for rationale selection and modify the rationale tokens necessary for the relation extraction task using relation labels. As a result, the unavailability of gold rationales can be addressed through end-to-end training. Our primary contributions include: (1) Introducing a novel end-to-end training system, {\modelname}, which treats rationale extraction as an adjustable binary mask for the relation extraction task and retains relevant, noise-free rationales via continuity and sparsity factors. (2) Experiments on four commonly used datasets demonstrate that {\modelname} significantly improves best-reported baselines in both full data and low-resource settings.

%% file: subfiles/4_model.tex
\section{Proposed Model}
\label{model}


\begin{figure}
    \centering
    \includegraphics[scale=0.29]{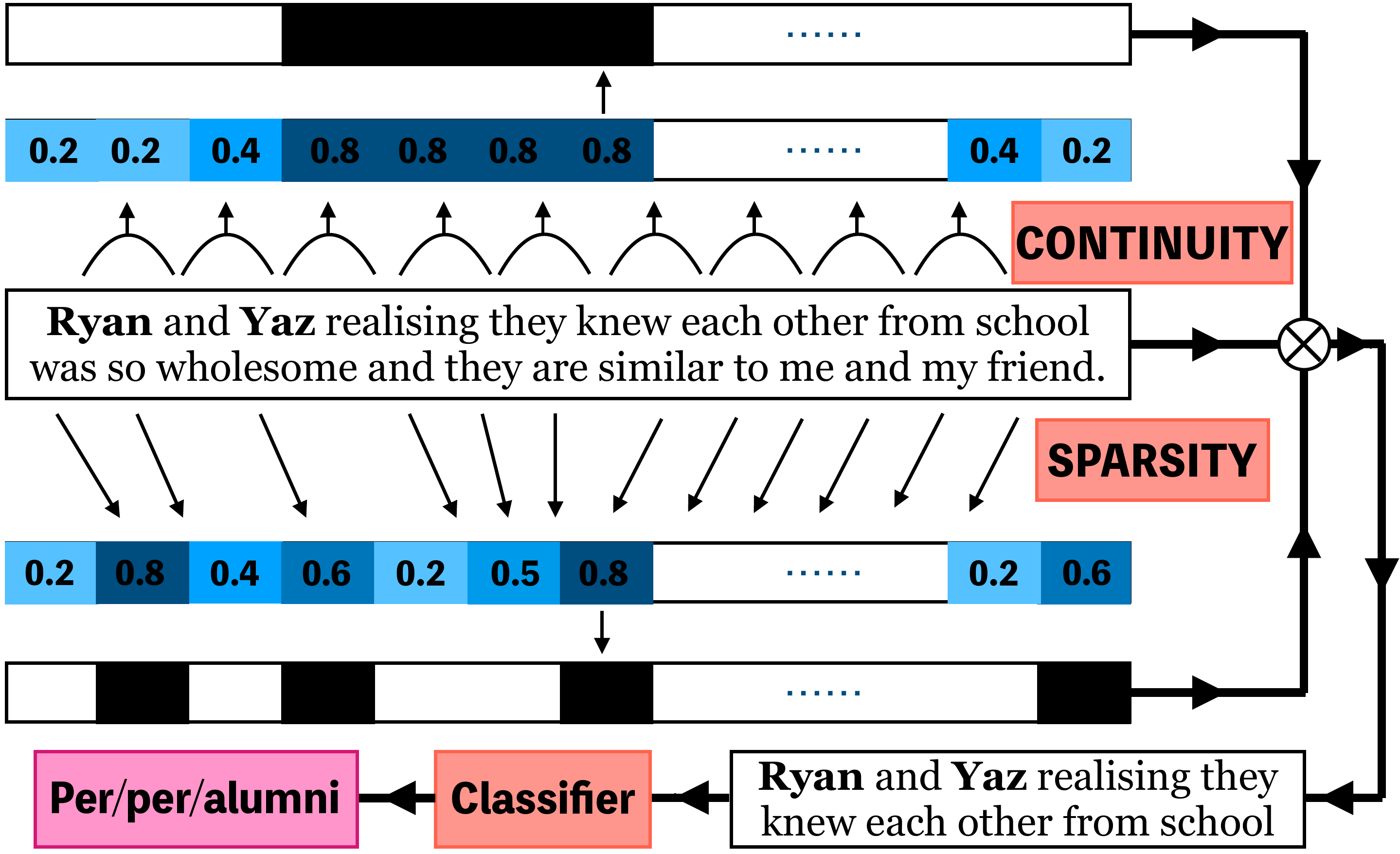}
    \vspace{-4mm}
    \caption{Architecture: The rationale extractor obtains the rationales from the input using a binary mask consists of continuity and sparsity factors.}
    \label{fig:overview}
    \vspace{-6mm}
\end{figure}

\subsection{Continuous Rationale Extractor}
In the continuous rationale extractor module of the model, we mask the tokens that are irrelevant to the relation extraction task, and keep the continuous tokens to improve extraction performance. 

\subsubsection{Sentence Representation and Importance Matrix}\label{Importance} We can obtain semantic embedding of each token through its contextualized sentence representation. In practice, we adopt the BERT \cite{devlin2019bert} to encode the token representations as: $\boldsymbol{x_{sent}} \in \mathbb{R}^{D\times L}$, where $D$ is the dimension of the embedding and $L$ is the number of tokens in the sentence. To select the tokens most relevant to the entities in the sentence for relation extraction task, we first calculate the importance matrix: $\boldsymbol{s} = \boldsymbol{x_{sent}}^\top \left(\boldsymbol{x_{e_1}} + \boldsymbol{x_{e_2}}\right)$ with the token embeddings of the two entities $\boldsymbol{x_{e_1}}$ and $\boldsymbol{x_{e_2}}$ extracted from $\boldsymbol{x_{sent}}$. We denote $\boldsymbol{s} = (s_1, ..., s_L)^\top$ and  obtain the importance score $s_i$ which represents the importance of the $i^{th}$ token towards RE task.

\subsubsection{Factor Graph}

We represent the token selection using a binary vector $\boldsymbol{m} = (m_1, m_2, ..., m_L)^{\top}$, where $m_i \in \{0,1\}$. The value 0 or 1 is used to indicate whether the $i^{th}$ token is selected. In this way, we could transform the structured prediction problem of rational sequence generation into the assignment of values to multiple variables. To maintain semantic coherence in token selection, it's important to consider the continuity in token choice. Additionally, to emphasize the importance and relevance of tokens to entities, we need to limit the number of tokens with sparsity. Therefore, we introduce the factor graph $\mathcal{F}$ and decompose these requirements into multiple local factors for optimal token selection. More specifically, we adopt the pairwise factor \texttt{CONTINUITY} and L-ary factor \texttt{SPARSITY}. In the following sections, we will formulate these two factors and  provide their score functions.
    
\texttt{CONTINUITY (CON)}: 
To improve the continuity of tokens selected for RE task, we adopt the \texttt{CONTINUITY (CON)} factor, which could examine whether each pair of consecutive tokens are both selected. We adopt the factor \texttt{CON} $(m_i, m_i+1; r_{i,i+1})$ to represent the constraint on the continuous selection of the $i^{th}$ and $(i+1)^{th}$ tokens. As shown in Figure \ref{fig:overview}, if both tokens are selected, we encourage this continuous selection by adding the edge score $r_{i,i+1} \geq 0$ in the score function. Formally, the score function for \texttt{CON} factor can be denoted as:
\begin{align}
        \mathrm{score}_{\texttt{CON}} (m_i, m_{i+1};r_{i,i+1}) = 
        m_i m_{i+1} r_{i,i+1}.
\end{align}

As illustrated in Section \ref{Importance}, we adopt the importance matrix $\boldsymbol{s}$ to measure the tokens that are relevant to the entities, which is a critical metrics to token selection.
We add the scores of the selected $i^{th}$ and $(i+1)^{th}$ tokens in the score function
and finalize the score function as:
\begin{align}
        \mathrm{score}_{\texttt{CON}} (m_i, m_{i+1};r_{i,i+1}) = 
        m_i m_{i+1} r_{i,i+1} + m_i s_i + m_{i+1} s_{i+1}.
\end{align}

We can impose continuity constraints on the token selection for the original sentence by leveraging the combination of the pairwise factors. Formally, 
the factor graph can be formulated as:
\begin{align}
        \mathcal{F} = \{\texttt{CON} (m_i, m_{i+1};r_{i,i+1}): 1 \leq i < L\}.
\end{align}

\texttt{SPARSITY (SPA)}: To control the sparsity in token selection for RE task, we adopt the L-ary factor \texttt{SPARSITY (SPA)} by imposing a limit $K$ on the maximum number of selected tokens as a restriction. In practice, $K$ can also be the proportion of all tokens in a sentence. The \texttt{SPA} factor is a hard constraint by definition. We can formulate the \texttt{SPA} factor with the following score function:
\begin{align} 
        \mathrm{score}_{\texttt{SPA}} (m_1, m_2, \cdots, m_L, K) = 
        \begin{cases}
        0, &\sum m_i \leq K,\\
        -\infty, &\sum m_i > K.
        \end{cases}
\end{align}
    
Overall, to consider continuity and sparsity together, we obtain the factor graph $\mathcal{F}$ by instantiating with the $L$ binary variables and combining the \texttt{CON} and \texttt{SPA} factors:
\begin{align}
        \mathcal{F} = 
            \{ \texttt{SPA} (m_1, ..., m_L; K) \} \cup
            \{ \texttt{CON}(m_i, m_{i+1}; r_{i,i+1}).
\end{align}

To utilize the both continuity and sparsity constraints and find an optimal solution for token selection, the score functions of factor graph $\mathcal{F}$ need to sum the local sub-problems $\{\texttt{CON} (m_i, m_{i+1};r_{i,i+1}):1 \leq i < L\}$ and $\{ \texttt{SPA} (m_1, ..., m_L; K) \}$ as:
\begin{align}
    \label{sum}
    \mathrm{score}(m; s) = 
    \begin{cases}
        \sum\limits_{i=1}^L m_i s_i 
        + \sum\limits_{i=1}^{L - 1} m_i m_{i+1} r_{i,i+1}, &\sum m_i \leq K,\\
        -\infty, &\sum m_i > K.
    \end{cases}
\end{align}
    
The hard constraint of $\mathcal{F}$ is inherited from the \texttt{SPA} factor, which specify that the total number of selected tokens should not exceed $K$. The soft constraint is inherited from the \texttt{CON} factors, encouraging to select consecutive tokens that are relevant to the RE task. To find a solution that satisfies these constraints well, we approach the problem of solving the variables as a Maximum A Posteriori (MAP) inference problem that maximizes the score function: score $(\boldsymbol{m}; \boldsymbol{s})$. We can represent this problem as maximization of the score function under the constraint that $|\boldsymbol{m}|_1 \leq K$:
\begin{align}
    \label{map}
    \hat{\boldsymbol{m}}=\arg \max _{|\boldsymbol{m}|_1\leq K, \boldsymbol{m} \in\{0,1\}^{L}}\underbrace{\big(\boldsymbol{s}^{\top} \boldsymbol{m}+\sum\limits_{i=1}^{L - 1} m_i m_{i+1} r_{i,i+1}\big)}_{\operatorname{score}(\boldsymbol{m} ; \boldsymbol{s})}.
\end{align}

In fact, maximizing the score function is essentially a complex structured problem involving sub-problems with interrelated global agreement constraints, making it difficult to find an accurate maximization algorithm \cite{NiculaeM20}. To solve this problem, we consider the Marginal Inference with Lagrange Multiplier.

\subsubsection{Marginal Inference with Lagrange Multiplier} 

We can solve the maximization problem with the Gibbs distribution and get an approximate solution. We construct a Gibbs distribution so that $p(\boldsymbol{m}; \boldsymbol{s}) \propto exp(\mathrm{score}(\boldsymbol{m}; \boldsymbol{s}))$. In this way, we can sample from $\hat{\boldsymbol{m}} \sim p(\boldsymbol{m}; \boldsymbol{s})$ and obtain an approximate optimal solution. However, obtaining unbiased samples is challenging. To address this issue, we use Perturb-and-MAP \cite{papandreou2011perturb}, an approximate sampling strategy.

Another problem is that the score function in Eq.~\ref{sum} is a piecewise function, making the Gibbs distribution $p(\boldsymbol{m}; \boldsymbol{s}) \propto exp(\mathrm{score}(\boldsymbol{m}; \boldsymbol{s}))$  discontinuous. As marginal inference in discontinuous Markov Random Fields is hard to solve, we reformulate the hard constraint: \texttt{SPA} in Eq.~\ref{sum} with Lagrange multiplier, which express hard constraints in the form of continuous functions. Specifically, we use a Lagrange Multiplier $\lambda > 0$, and add $\lambda(K - |\boldsymbol{m}|_1)$ to the objective function in Eq.~\ref{map}. 
We finalize the Eq.~\ref{map} as:
\begin{align}
        \hat{\boldsymbol{m}}=\arg \max _{|\boldsymbol{m}|_1\leq K, \boldsymbol{m} \in[0,1]^{L}} \left(\boldsymbol{s}^{\top} \boldsymbol{m}
        + \sum\limits_{i=1}^{L - 1} m_i m_{i+1} r_{i,i+1} + \lambda(K - |\boldsymbol{m}|_1)\right),
\end{align}
where the Gibbs distribution should be reformulated as $p(\boldsymbol{m}; \boldsymbol{s}) \propto exp(\mathrm{score}(\boldsymbol{m}; \boldsymbol{s}) + \lambda(K - |\boldsymbol{m}|_1))$. Therefore, the reformulated Gibbs distribution becomes continuous, enabling us to calculate the optimal $\boldsymbol{m}$ that maximizes the score function, and obtain the rationales which are relevant to the relation extraction task.



    

\subsection{Relation Classifier}
Finally, the classifier makes relation predictions conditioned on the selected rationales and the entities $\boldsymbol{e}$: $\hat{\boldsymbol{y}} = \texttt{pred}(\boldsymbol{m}\odot\boldsymbol{x}\mathbin\Vert\boldsymbol{e})$ to obtain the relation label distributions. $\odot$ and $\mathbin\Vert$ denote the element-wise product and concatenation, respectively. The relation classification loss could be calculated as: $\mathcal{L}=-\sum_{i=1}^{N}\boldsymbol{y}_{i}\log\hat{\boldsymbol{y}}_{i}$,
where $N$ is the number of training sentences in an epoch, and $\boldsymbol{y}_{i}$ is the ground-truth tag vector of the sentence $x_{i}$. The relation classification loss could jointly train the Continuous Rationale Extractor module and Relation Classifier module in an end-to-end manner.

\input{tables/results.tex}

%% file: tables/results.tex
\begin{table*}[t]
\centering
\caption{Average micro F1 results in four RE datasets. ``re.'' means that we will replace the entity information verbalization in SURE with the corresponding entity thinking baselines. We mark the (standard deviation) of the results.}
\vspace{-3mm}
\scalebox{0.67}{
\begin{tabular}{lcccccccccccccccc}
\thickhline
\multicolumn{1}{c}{\multirow{2}{*}{Methods/Datasets}}  &\multicolumn{4}{c}{SemEval}&  \multicolumn{4}{c}{TACRED} &   \multicolumn{4}{c}{TACRED-Revisit} &   \multicolumn{4}{c}{Re-TACRED}  
\\ \cmidrule(lr){2-5} \cmidrule(lr){6-9} \cmidrule(lr){10-13} \cmidrule(lr){14-17}
&10\% & 25\% & 50\% & 100\%& 10\% & 25\% & 50\% & 100\% &  10\% & 25\% & 50\% & 100\% &  10\% & 25\% & 50\% & 100\%  \\
\midrule 
\textbf{SURE}  & 77.2 & 81.5 & 83.9 & 86.3 & 67.9 & 70.4 & 71.9 & 73.3  & 72.3 & 75.1 & 77.4 & 79.2 & 78.5 &  82.6 &84.7  & 88.2 \\
\cmidrule(lr){2-17}
\multicolumn{1}{l}{re. MTB \cite{soares2019matching}} &76.3	&80.8 &	82.9	&85.4&	67.2	&69.8	&71.1	&72.3&	71.4&	74.3&	76.6&	78.4&	77.6&	81.8&	83.8	&87.2 \\ 
\multicolumn{1}{l}{re. Entity Mask \cite{zhang2017position}}& 76.7	&80.9&	83.3&	86.0&	67.3&	70.0	&71.4	&72.5	&71.8	&74.7&	76.9	&78.5	&77.8&	82.2&	84.3&	87.6 \\ 
\multicolumn{1}{l}{re. Typed Marker \cite{zhou2022improved}} & 77.0&	81.3&	83.7	&86.3	&67.8	&70.4&	71.8&	73.2&	72.1&	75.0&	77.2&	79.0&	78.2&	82.4	&84.5&	88.1  \\
\cmidrule(lr){2-17}
\multicolumn{1}{l}{+CFIE \cite{nan2021uncovering}} &77.3	&81.7&	84.1&	86.6&	68.2&	70.5	&72.2	&73.5	&72.5&	75.2&	77.6	&79.4&	78.6&	82.8&	85&	88.4\\ 
\multicolumn{1}{l}{+CORSAIR \cite{qian2021counterfactual}}& 77.4&	81.8	&84.1&	86.7&	68.3&	70.7	&72.4&	73.7	&72.7	&75.4&	77.7	&79.6&	78.9&	83.1	&85.1&	88.6 \\
\multicolumn{1}{l}{+CORE \cite{wang2022should}}& 77.5&	82.0	&84.2	&86.7&	68.3&	70.8	&72.3&	73.7	&72.8	&75.6&	77.8&	79.6&	78.9&	83.2&	85.3&	88.7 \\
\cmidrule(lr){2-17}
\multicolumn{1}{l}{+HardKuma \cite{bastings2019interpretable}} &77.4	&81.4&	84.0	&86.5&	68.0&	70.6	&72.2&	73.5	&72.5&	75.3&	77.6&	79.4	&78.7&	82.7&	85.0&	88.3 \\
\multicolumn{1}{l}{+IB Objective \cite{paranjape2020information}}&77.5&	81.7	&84.3&	86.6	&68.3	&70.7&	72.3&	73.6	&72.6&	75.3&	77.8	&79.5&	79.0&	83.0	&85.1&	88.6 \\
\multicolumn{1}{l}{+UNIREX \cite{chan2022unirex}}  &77.8&	81.9&	84.4&	86.7&	68.5&	70.9	&72.5&	73.8	&72.8	&75.7&	78.0&	79.7&	79.2&	83.1&	85.4	&88.8  \\

\cmidrule(lr){2-17}
\rowcolor{gray!15}
\multicolumn{1}{l}{+\textbf{{\modelname}}}  & \textbf{78.2\small{(0.2)}} & \textbf{82.3\small{(0.1)}} & \textbf{84.8\small{(0.1)}} & \textbf{87.2\small{(0.2)}}  &\textbf{68.9\small{(0.3)}} &\textbf{71.3\small{(0.2)}} &\textbf{73.0\small{(0.3)}} &\textbf{74.2\small{(0.1)}} &\textbf{73.3\small{(0.2)}} & \textbf{76.0\small{(0.2)}} & \textbf{78.4\small{(0.1)}} & \textbf{80.1\small{(0.1)}} &\textbf{79.6\small{(0.3)}}  &\textbf{83.6\small{(0.2)}}  & \textbf{85.7\small{(0.1)}} &\textbf{89.1\small{(0.2)}}  \\
\rowcolor{gray!15}
\multicolumn{1}{l}{\quad \textit{w/o Continuity}}  & 77.8\small{(0.3)}	 &81.9\small{(0.4)}	 &84.2\small{(0.2)}	&86.8\small{(0.4)}&	68.5\small{(0.3)}	&70.7\small{(0.2)}	&72.5\small{(0.3)}	&73.8\small{(0.2)}	&72.8\small{(0.4)}&	75.5\small{(0.3)}	&77.9\small{(0.3)}&	79.7\small{(0.2)}	&79.1\small{(0.4)}	&83.2\small{(0.3)}&	85.3\small{(0.2)}	&88.8\small{(0.3)} \\
\rowcolor{gray!15}
\multicolumn{1}{l}{\quad \textit{w/o Sparsity}} &77.6\small{(0.2)}	&81.8\small{(0.2)}	&84.3\small{(0.1)}&	86.6\small{(0.3)}&	68.4\small{(0.1)}	&70.5\small{(0.2)}	&72.4\small{(0.3)}	&73.6\small{(0.1)}&	72.7\small{(0.2)}	&75.5\small{(0.3)}&	77.8\small{(0.3)}	&79.8\small{(0.2)}&	79.0\small{(0.2)}&	83.1\small{(0.3)}	&85.1\small{(0.1)}&	88.6\small{(0.1)} \\
\rowcolor{gray!15}
\multicolumn{1}{l}{\quad \textit{w/o Adding Entities}} &78.0\small{(0.2)}	&82.2\small{(0.1)}	&84.6\small{(0.2)}&	87.1\small{(0.2)}&	68.8\small{(0.3)}	&71.1\small{(0.2)}	&72.8\small{(0.2)}	&74.0\small{(0.1)}	&73.1\small{(0.2)}	&75.9\small{(0.2)}&	78.2\small{(0.1)}	&80.0\small{(0.1)}&	79.3\small{(0.3)}	&83.4\small{(0.2)}&	85.6\small{(0.2)}&	89.0\small{(0.1)} \\

\thickhline
\end{tabular}}
\label{tab:verification}
\vspace{-3mm}
\end{table*}

%% file: subfiles/5_experiment.tex
\section{Experiments and Analyses}
\label{sec:experiments}

\subsection{Experimental Setup and Baselines}
\noindent \textbf{Setup:}
We evaluate the model on four widely-used RE datasets: SemEval \cite{hendrickx2010semeval}, which contains 6,507/1,493/2,717 samples in train/dev/test sets and 19 relation types. TACRED \cite{zhang2017position} and TACRED-Revisit \cite{alt2020tacred}, which contain 68,124/22,631/15,509 samples and 42 relation types. Re-TACRED \cite{stoica2021re}, which contains 58,465/19,584/13,418 samples and 42 relation types. Following prior effort \cite{wang2022should}, we adopt Micro F1 as the evaluation metric. Under the low-resource setting, we randomly sample 10\%, 25\%, and 50\% of the training set as the small-scale training sets for evaluation, and evaluate our model on the test set. We use the BERT-Base default tokenizer with a max-length of 128 to preprocess data. We set K as 60\% of all tokens in the sentence. For the classifier, we set the layer dimensions as $768$-${384}$-labels. We use BertAdam~\citep{kingma2014adam} with 3e-5 learning rate, warm up with 0.06 to optimize the loss and set the batch size as 16.

\noindent \textbf{Baselines:}
We first introduce SOTA models as base model on the RE task, and then adopt various baselines. We adopt SURE \cite{lu2022summarization} as the base model. We compare {\modelname} with the following baselines: \textit{Entity thinking} baselines: (1) MTB \cite{soares2019matching}, (2) Entity Mask \cite{zhang2017position}, (3) Typed Entity Marker \cite{zhou2022improved}. \textit{Counterfactual thinking} baselines adopt causal inference to remove bias in RE tasks: (4) CFIE \cite{nan2021uncovering}, (5) CORSAIR \cite{qian2021counterfactual}  (6) CORE \cite{wang2022should}. \textit{Rationale thinking} baselines could predict sparse binary masks over input tokens for RE tasks: (7) HardKuma \cite{bastings2019interpretable}, (8) IB objective \cite{paranjape2020information}, (9) UNIREX \cite{chan2022unirex}. Note that the entity thinking method is also used in the SURE, all baselines of entity thinking are used to replace the methods in SURE.

\subsection{Results and Analysis}

\noindent \textbf{Overall Performance.}
Table \ref{tab:verification} shows the mean and standard deviation results with 5 runs of training and testing on four datasets. We observe that using the entity information verbalization (SURE \cite{lu2022summarization}) can achieve an average 0.6\% improvement in F1 across all datasets compared to other entity thinking methods. Therefore, we adopt SURE as the base model. For counterfactual and rationale thinkings, we find that they both bring a 0.3\% improvement in the F1 performance across all datasets. Our proposed method of rational thinking addresses two major challenges: (1) end-to-end training of both the rationale extraction and relation classifier, and (2) extraction of continuous rationales. As a result, {\modelname} achieves a significant 0.9\% improvement in F1 across all RE datasets, including low-resource RE settings. Compared with the previous SOTA: UNIREX, {\modelname} has an additional 0.4\% increase in F1 performance. An interesting finding is that for low-resource settings (e.g., only 10\% of the training set), {\modelname} can achieve more performance improvements than the full data setting: 1.1\% vs. 0.9\%, which shows that {\modelname} is robust enough in the case of limited training data. The low-noise and relevant rationales obtained by using {\modelname} can help the F1 performance of the base model more significantly. 

\begin{figure}
    \centering
    \includegraphics[scale=0.31]{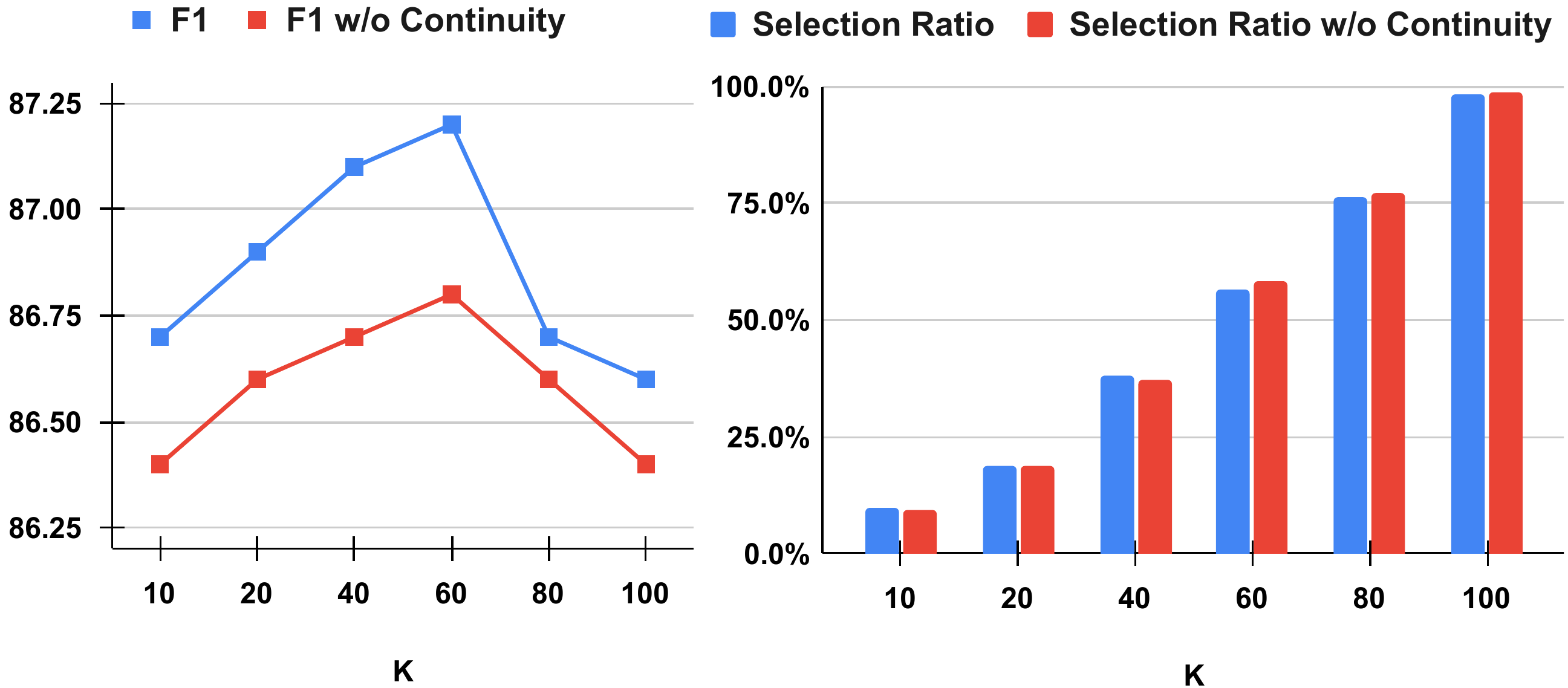}
    \vspace{-4mm}
    \caption{Effect of Two Factors (Continuity and Sparsity). $K$ is the hyper-parameter to control the sparsity of the token selection. Continuity is imposed to improve contiguity. }
    \label{fig:factorK}
    \vspace{-5mm}
\end{figure}

\noindent \textbf{Ablation Study.}
We perform an ablation study to demonstrate the effectiveness of our model's various modules on the test set. {\modelname} \textit{without Continuity} and {\modelname} \textit{without Sparsity} eliminate the continuity and sparsity elements in the factor graphs in rationale extraction, respectively. {\modelname} \textit{without Adding Entities} removes the entities added in the relation classifier module, using only the rationales for relation classification. Table \ref{tab:verification} generally concludes that all modules positively impact performance. Specifically, the absence of continuity leads to discontinuous rationales, affecting the coherence of semantic representations and causing a 0.4\% F1 performance drop. Removing sparsity selects noisier rationales, resulting in a 0.5\% F1 performance reduction. Interestingly, removing added entities has minimal effect on F1 performance (0.1\%). We find that 89\% of rationales contain two entities and 97\% contain at least one entity, indicating that adding entities provides little additional information.

\noindent \textbf{Effect of Two Factors:}
As shown in Figure~\ref{fig:factorK}, we display F1 scores and token selection rates in relation to varying $K$ values on  SemEval. As $K$ rises, more tokens within sentences are chosen as rationales. Nonetheless, the F1 score for {\modelname} doesn't increase consistently with higher $K$, due to the incorporation of unrelated rationales. Optimal performance occurs at $K=60$, meaning 60\% of tokens are selected as rationales on average. Eliminating the Sparsity factor entirely causes the model's F1 score to decline from 87.2 to 86.6. Additionally, the Continuity constraint benefits the model, as {\modelname} with Continuity constraints consistently produces improved outcomes.

\noindent \textbf{Coherence Analysis of Rationales.}
{\modelname} utilizes the continuity factor to control the generation of rationales that are more semantically coherent, which can express more fluent semantics. We analyze the coherence of the rationales through perplexity based on GPT-3 \cite{radford2019language}. From Table \ref{tab:perplexity}, {\modelname} could obtain the lowest average perplexity, approaching that of the original sentences.

\noindent \textbf{Human Evaluation.}
We conduct human evaluations of rationales with a 15-member annotation team, involving 5 members in data validation. Annotators predict relation labels using original sentences and extracted rationales, then rate information sufficiency (on a 1-5 scale) for both. Higher scores signify greater sufficiency. To ensure consistency, we perform inter-annotator agreement and manual validation. Table \ref{tab:human} shows that annotators can provide more accurate relation labels even with lower information sufficiency in rationales than original sentences, suggesting that removing irrelevant details from sentences can decrease noise and enhance relational prediction accuracy.

\input{tables/corhence.tex}
\input{tables/human.tex}

%% file: tables/corhence.tex
\begin{table}[t!]
\centering
\caption{Perplexity of the extracted rationales. Original means the original sentences. Lower perplexity is better.}
\vspace{-2mm}
\resizebox{0.85\linewidth}{!}{
\begin{tabular}{lcccc}
\thickhline
Methods / Datasets & SemEval & TACRED  & TACRED-Revisit   & Re-TACRED  \\
\midrule 
\multicolumn{1}{l}{HardKuma \cite{bastings2019interpretable}} &11.37  &13.35 &12.21 &11.93 \\
\multicolumn{1}{l}{IB Objective \cite{paranjape2020information}}& 10.37  &12.23 &11.56 &10.74\\
\multicolumn{1}{l}{UNIREX \cite{chan2022unirex}} & 13.42  &12.64 & 14.22 & 13.87\\
\rowcolor{gray!15}
\multicolumn{1}{l}{{\modelname}} & \textbf{5.13} &\textbf{5.68} &\textbf{6.02} &\textbf{5.75}  \\
\hline
\multicolumn{1}{l}{Original Sentences} & 3.75 &4.02 &3.83  &3.68\\
\thickhline
\end{tabular}}
\label{tab:perplexity}
\vspace{-2mm}
\end{table}

%% file: tables/human.tex
\begin{table}[t!]
\centering
\caption{Human evaluation (Micro F1 / Information Sufficiency) of the original sentences and extracted rationales.}
\vspace{-2mm}
\resizebox{0.9\linewidth}{!}{
\begin{tabular}{lcccc}
\thickhline
Datasets & SemEval & TACRED  & TACRED-Revisit   & Re-TACRED  \\
\midrule 
\rowcolor{gray!15}
\multicolumn{1}{l}{Extracted Rationales} & 95.5 / 4.4 & 89.3 / 4.0 & 93.5 / 4.2 &  96.8 / 4.5   \\
\hline
\multicolumn{1}{l}{Original Sentences} & 94.3 / 4.5 & 87.6 / 4.3  & 92.3 / 4.2 &  96.0 / 4.6\\
\thickhline
\end{tabular}}
\label{tab:human}
\vspace{-2mm}
\end{table}

%% file: subfiles/6_conclusion.tex
\section{Conclusion and Future Work}
\label{conclusion}

In this paper, we propose a novel rationale extraction framework {\modelname}, which adopt two factors, continuity and sparsity, to control the relevancy of rationales to the RE task and improve the coherence. We introduce the marginal inference with a Lagrange multiplier to solve the problem of maximizing the score function with two factors. Therefore, we could jointly train the rationale extraction and relation classification tasks in an end-to-end manner where gold annotations for rationales are not available. Experiments on four datasets show the effectiveness of {\modelname}. In the future, we can extend the research on relation extraction to the construction of knowledge graphs \cite{fang2017effective, zhang2022knowledge, lkgr, chen2022attentive}, the matching of knowledge graphs \cite{ClusterEA, DualMatch, LargeEA22, EASY21}, and the acceleration of information retrieval \cite{bgch, bgr}.